\documentclass{article}
\usepackage{spconf,amsmath,graphicx}
\usepackage{subcaption}

\usepackage{enumitem}
\setlist{nosep, leftmargin=14pt}

\usepackage{mwe} % to get dummy images

\title{Automated Prediction of Breast Cancer Response to Neoadjuvant Chemotherapy from DWI Data}
\name{Shir Nitzan$^{\star}$ \qquad Maya Gilad$^{\dagger}$ \qquad Moti Freiman$^{\dagger}$}

\address{$^{\star}$Efi Arazi School of Computer Science, Reichman University, Herzliya, Israel \\
$^{\dagger}$Faculty of Biomedical Engineering, Technion – Israel Institute of Technology, Haifa, Israel}

\begin{document}
%\ninept
%
\maketitle
\begin{abstract}
Effective surgical planning for breast cancer hinges on accurately predicting pathological complete response (pCR) to neoadjuvant chemotherapy (NAC). Diffusion-weighted MRI (DWI) and machine learning offer a non-invasive approach for early pCR assessment. However, most machine-learning models require manual tumor segmentation, a cumbersome and error-prone task. We propose a deep learning model employing "Size-Adaptive Lesion Weighting" for automatic DWI tumor segmentation to enhance pCR prediction accuracy. Despite histopathological changes during NAC complicating DWI image segmentation, our model demonstrates robust performance. Utilizing the BMMR2 challenge dataset, it matches human experts in pCR prediction pre-NAC with an area under the curve (AUC) of 0.76 vs. 0.796, and surpasses standard automated methods mid-NAC, with an AUC of 0.729 vs. 0.654 and 0.576. Our approach represents a significant advancement in automating breast cancer treatment planning, enabling more reliable pCR predictions without manual segmentation.

% Surgical planning for breast cancer greatly depends on accurate prediction of the pathological complete response (pCR) to neoadjuvant chemotherapy (NAC). The advent of diffusion-weighted MRI (DWI) combined with machine-learning methods presents a promising avenue for non-invasive early assessments of pCR. Yet, existing machine-learning methods using DWI data for pCR prediction often require manual tumor segmentation. A process that is both tedious and prone to errors. Deep-learning-based automatic segmentation holds the promise to fully automate the pCR prediction process. Nonetheless, segmenting DWI images obtained during NAC remains a challenge due to treatment-induced histopathological changes. We introduced a deep learning model with a ``Size-Adaptive Lesion Weighting'' loss to overcome these challenges, automating DWI tumor segmentation and enhancing the accuracy of radiomics-based pCR predictions from DWI data. Employing the BMMR2 challenge dataset, our algorithm matched the proficiency of human experts in predicting pCR prior to NAC with an area under the curve (AUC) of 0.76 VS. 0.796. Furthermore, in forecasting pCR midway through NAC, our model exhibited enhanced AUC performance relative to automated segmentation methods that did not incorporate Weight Lesion enhancement, achieving 0.729 compared to 0.654 and 0.576. Our proposed model offers a promising step forward in optimizing breast cancer treatment plans without the need for manual intervention.
\end{abstract}
\section{Introduction}
\label{sec:intro}

Breast cancer ranks among the most common cancers, causing significant mortality among women worldwide \cite{bhushan2021current}. The introduction of Neoadjuvant chemotherapy (NAC) has significantly improved survival outcomes by effectively treating invasive breast cancer \cite{von2012definition}. The pathological complete response (pCR), indicating the complete eradication of invasive disease in the breast or axillary lymph nodes, serves as a pivotal metric for assessing NAC's success. Importantly, pCR informs treatment plans and helps predict patient outcomes \cite{partridge2018diffusion}.

Prompt pCR prediction is vital during treatment. The advent of multi-parametric MRI (mp-MRI) presents a promising avenue for these early assessments \cite{liang2022machine}. Methods like dynamic contrast-enhanced MRI (DCE-MRI) and diffusion-weighted MRI (DWI) are particularly noteworthy for their enhanced sensitivity to tissue micro-structure, providing an edge over traditional anatomical MRI techniques \cite{partridge2018diffusion, Baltzer2020}. Recently, the adoption of radiomics and high-dimensional imaging characteristics has become increasingly popular in forecasting pCR after NAC, leveraging mp-MRI data from breast cancer patients \cite{gilad2022pd, joo2021multimodal}.

The use of tumor segmentation in radiomics-based pCR prediction with DWI data poses challenges for clinical application. Traditional manual segmentation, while widely adopted, is labor-intensive and prone to significant variability between operators \cite{partridge2018diffusion, Timmeren2020}. This variability can impact the consistency of pCR predictions. Automated image segmentation emerges as a preferred solution, minimizing intra- and inter-observer discrepancies in radiomics features \cite{Timmeren2020}. Nonetheless, segmenting DWI images obtained during NAC remains a challenge due to treatment-induced histopathological changes. NAC can provoke diverse changes in tumor cellularity, from concentric shrinkage patterns to the disintegration into isolated tumor cell clusters.
% \cite{goorts2018mri}.

\begin{figure*}[t!]
  \centering
  \includegraphics[width=0.87\textwidth]{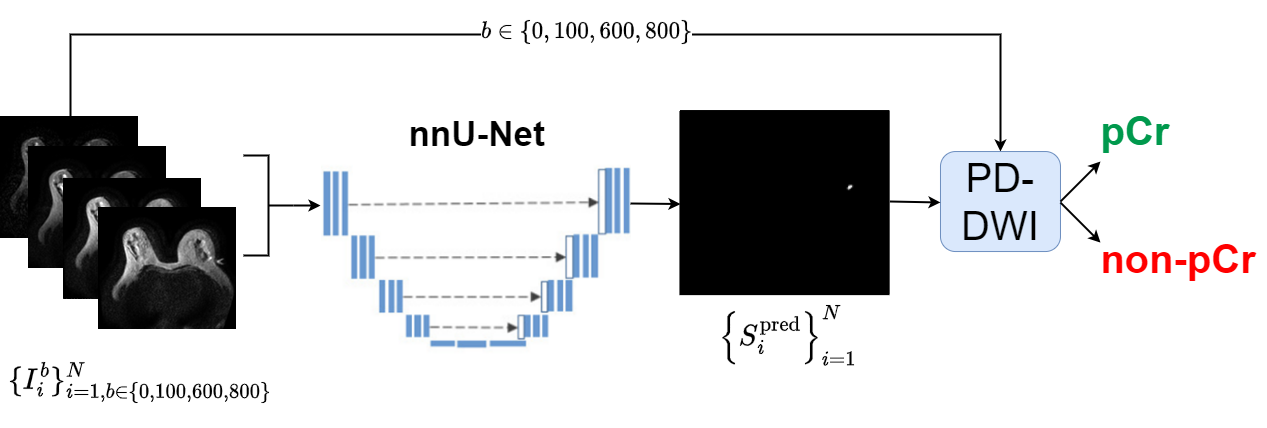}
  \caption{Automated Workflow for pCR Prediction: DWI data is processed by the nnU-Net for tumor segmentation using a tailored loss function. The resulting segmentation, combined with DWI, informs the PD-DWI model to forecast pCR.
  % Automated pCR Prediction Workflow: DWI data inputs the nnU-Net for tumor segmentation with a custom loss. Predicted segmentation and DWI then drive the PD-DWI model for pCR prediction.
  }
  \label{workflow}
\end{figure*}

In this study, we present the ``Size-Adaptive Lesion Weighting'' method, a specialized weighted loss function crafted to tackle data imbalance, particularly evident when tumors undergo significant shrinkage during NAC. This ensures an equitable lesion representation at the voxel level throughout treatment. Building on this loss function, we integrated it into the nnU-Net architecture \cite{isensee2021nnu}, refining tumor segmentation in DWI data. This enhanced pipeline, with its focus on accurate segmentation, paves the way for the subsequent application of the PD-DWI model \cite{gilad2022pd}, which leverages both the DWI data and the predicted segmented lesions to forecast treatment outcomes. Fig. \ref{workflow} visualized the proposed pipeline for fully automatic pCR prediction from DWI data.

\section{METHOD}
\subsection{Data}
We utilized the BMMR2 challenge dataset, sourced from the ACRIN 6698 multi-center study \cite{clark2013cancer, partridge2018diffusion, li2024breast}. This dataset encompasses 191 participants from diverse institutions. The challenge organizers pre-divided it into training and testing subsets (60\%-40\% distribution, stratified by pCR results). Every participant underwent sequential mp-MRI exams, comprising standardized DWI and DCE-MRI scans at three intervals: prior to NAC (T0), 3 weeks post-initiation (T1), and 12 weeks after (T2). Diffusion gradients were three-dimensional with b-values of 0, 100, 600, and 800 s/mm$^2$, without employing respiratory triggers or motion compensation techniques \cite{partridge2018diffusion}. The dataset offered manual DWI whole-tumor and DCE-MRI functional tumor segmentation for each interval, alongside complete ADC and Signal Enhancement Ratio (SER) maps. Supplementary non-imaging data covered demographics (age, race), 4-level lesion type, 4-level hormonal receptor HR/HER2 status, 3-level tumor grade, and MRI-measured tumor diameter at T0.

\subsection{Fully Automatic pCR Prediction}

The flow and interactions between the two main stages of our pipeline can be seen in Fig. \ref{workflow}.

\subsubsection{Segmentation}
MRI segmentation involves dividing an image into its constituent anatomical regions of interest, formally represented as:
% MRI segmentation typically involves partitioning a given image into its constituent anatomical regions of interest. In the realm of deep learning, the segmentation task can be formalized as:
\begin{equation}
S_{\theta}(x) = \hat{y}
\end{equation}
where \( x \) denotes the input image, \( S_{\theta} \) represents the segmentation network function with parameters \( \theta \), and \( \hat{y} \) is the predicted segmentation map.

The parameters \( \theta \) are learned by minimizing a loss function that measures the difference between the predicted segmentation \( \hat{y} \) and the ground truth labels \( y \):
\begin{equation}
\hat{\theta} = \arg\min_{\theta} \sum_{i=0}^{n_{\text{data}}} \mathcal{L}(S_{\theta}(x^{(i)}), y^{(i)})
\end{equation}
where \( \mathcal{L} \) denotes a suitable loss function, \( n_{\text{data}} \) is the number of training examples, \( x^{(i)} \) represents the \( i \)-th image, and \( y^{(i)} \) is its corresponding ground truth segmentation.

% Dice Loss: A Common Choice for Medical Image Segmentation:
The Dice coefficient is a classical benchmark for evaluating segmentation, by quantifying the similarity between the segmentation maps. It has been adopted as a loss function, known as Dice Loss, to optimize neural network training in medical imaging.

However, the Dice loss may not perform well on imbalanced data due to the equal weighting of false positives (FP) and false negatives (FN). This limitation has led to the adoption of the Tversky loss which accounts for class imbalance by introducing different weights for FP and FN with parameters \(\alpha\) and \(\beta\) as follows:
\begin{equation}
T L(p, \hat{p}) = 1 - \frac{1 + p \hat{p}}{1 + p \hat{p} + \alpha(1 - p)\hat{p} + \beta p(1 - \hat{p})}
\end{equation}
However, Tversky loss also presents its own limitations, as it operates at a voxel level, neglecting the scale of lesions entirely during the learning process. Consequently, this leads to a model bias wherein voxels from larger lesions are overrepresented in the learning phase compared to those from smaller lesions, whose volume is relatively negligible. 
Such a disparity in learning result in suboptimal performance when detecting smaller lesions, which are critical in early disease diagnosis and treatment monitoring. The Universal Loss Reweighting \cite{shirokikh2020universal} aimed to prioritize smaller lesions but was limited by its linear approach, not fully addressing the lesion size range.

To address these limitations, we present a "Size-Adaptive Lesion Weighting" loss function, inspired by \cite{fenneteau2022size}. Our method, focusing on exponential weighting, finely tunes voxel weights based on lesion size to prioritize small lesion detection within convolutional neural networks. 

% This strategy ensures equitable lesion representation across sizes.

% , addressing previous linear approach disparities and enhancing detection accuracy comprehensively.

% To address this limitation, we introduce a "Size-Adaptive Lesion Weighting" loss function, inspired by the methodology proposed in \cite{fenneteau:hal-03836787}. This approach is designed to prioritize the importance of small lesions during the training of convolutional neural networks by adjusting the voxel weights according to lesion size. It ensures a balanced representation across lesion sizes and mitigates the disparities in voxel-level lesion representation.

Our loss function leveraging weighted lesion strategy
and incorporated Tversky loss function to adjust the global balance between FPs and FNs and cross-entropy (CE) loss for optimal results.
Let $B, H, W, D$ represent the batch size, height, width, and depth, respectively. Let $P \in \{0,1\}^{B \times H \times W \times D}$ be the ground truth segmentation, and $\hat{P} \in [0,1]^{B \times H \times W \times D}$ denote the model predictions.

The lesion weight function \( \omega(v_{\text{les}}) \) is given as:
\begin{equation}
\omega(v_{\text{les}}) = w_{\text{max}} - \frac{(w_{\text{max}} - w_{\text{min}})}{1 + \alpha \cdot  e^{ \frac{-k v_{\text{les}}}{\text{vrange}}}}
\end{equation}

Where \( v_{\text{les}} \) is the lesion volume, \( w_{\text{max}} \) and \( w_{\text{min}} \) are the maximum and minimum weightings, respectively. \( \alpha \) adjust the x-axis translation of the curve, \textit{range} defines the \( v \) range between the curve's minimum
and maximum asymptotes, and \( k \) determines the curve's steepness.

% \begin{figure}[!b] 
%     \centering

%     \includegraphics[width=0.5\textwidth]{weight_function_plot.png} % Adjust the width as necessary
% \caption{Size-Adaptive Lesion Weighting function relating lesion volume, \( v_{\text{les}} \), to its weight, \( w(v_{\text{les}}) \).}

%     \label{fig:lesion_weight_function} 
% \end{figure}

For each voxel's coordinate \((b, h, w, d)\), we define:
\begin{align*}
TP_{b,h,w,d} &= P_{b,h,w,d} \cdot \hat{P}_{b,h,w,d} \\
FN_{b,h,w,d} &= P_{b,h,w,d} \cdot (1 - \hat{P}_{b,h,w,d}) \\
FP_{b,h,w,d} &= (1 - P_{b,h,w,d}) \cdot \hat{P}_{b,h,w,d}
\end{align*}

We define the Weighted Lesion Tversky (WLT) loss function as:
\begin{equation}
\scriptsize
\text{L}_{\text{WLT}} = - \frac{\sum \text{TP}_{b,h,w,d} \cdot \Omega_{b,h,w,d}}{\sum \text{TP}_{b,h,w,d} + \alpha \cdot \sum \text{FP}_{b,h,w,d} + \beta \cdot \sum  \text{FN}_{b,h,w,d} \cdot \Omega_{b,h,w,d}}
\end{equation}
where:
\(\Omega_{b,h,w,d}\) is the weight map generated by associating each voxel with the value \(\omega(v_{\text{les}})\) where \(v_{\text{les}}\) is the lesion volume it belongs to.

% The Tversky loss parameters \(\alpha\) and \(\beta\) control the global magnitude of penalties for \(FPs\) and \(FN\) respectively.
The sums are taken over all \(b \in [1,B]\), \(h \in [1,H]\), \(w \in [1,W]\), and \(d \in [1,D]\).

This loss function encourages the accurate identification of smaller lesions by applying the lesion weight function at each voxel to \({FN}_{b,h,w,d}\), while controlling the global weight of the \(FP\) term with a \(\alpha\) parameter.

\begin{figure*}
    \centering
% Titles
\begin{tabular}{cccc}
    \multicolumn{1}{p{0.24\textwidth}}{\centering Reference Manual} & 
    \multicolumn{1}{p{0.24\textwidth}}{\centering \hspace{-2em}Lesion Weighting} & 
    \multicolumn{1}{p{0.24\textwidth}}{\centering \hspace{-3em}CE with Tversky} & 
    \multicolumn{1}{p{0.24\textwidth}}{\centering \hspace{-5em}Tversky}

\end{tabular}
    
  \vspace{1em} % Space between titles and images
    % Second Row
    \begin{subfigure}{0.22\textwidth}
        \includegraphics[width=\linewidth]{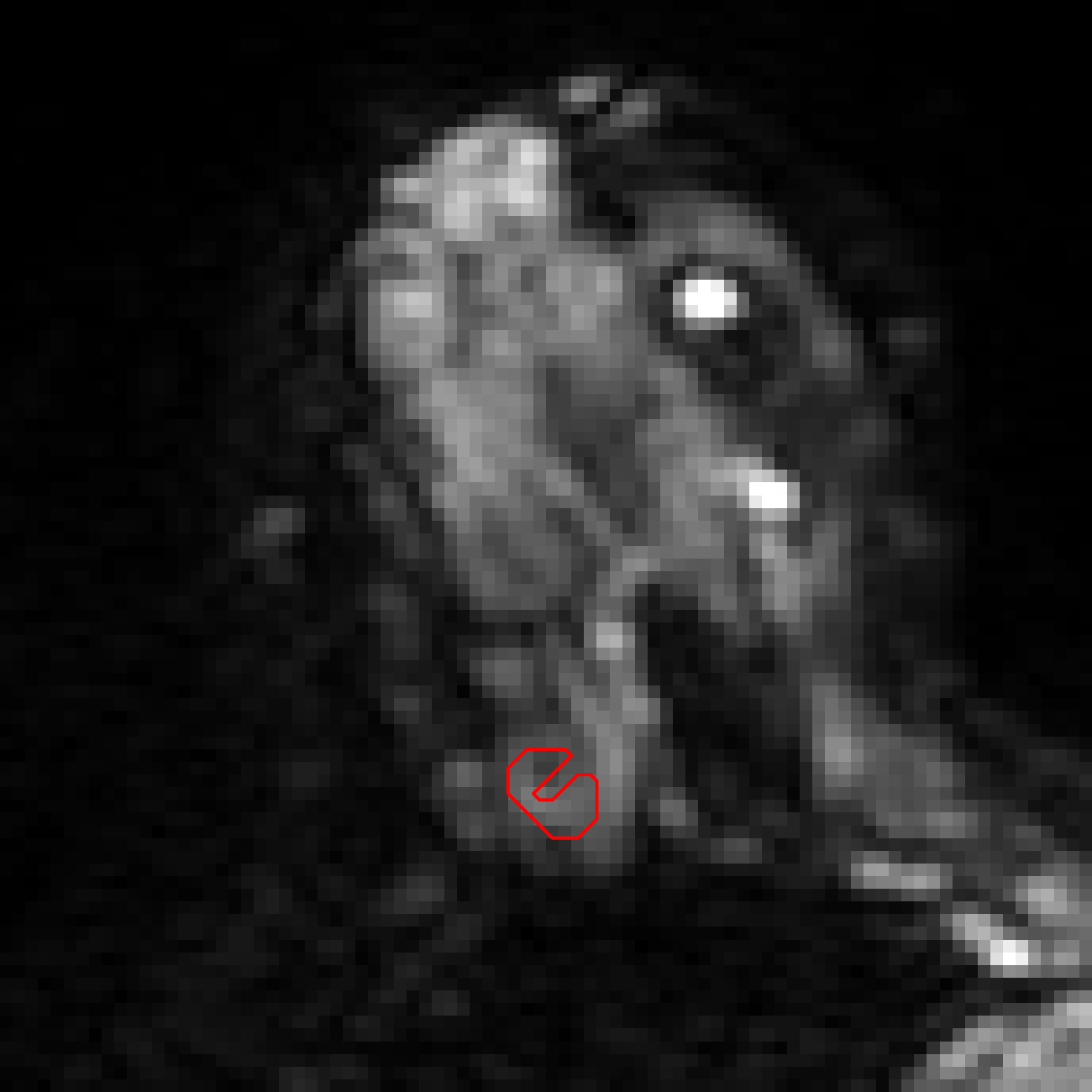}
    \end{subfigure}
    \hfill
    \begin{subfigure}{0.22\textwidth}
        \includegraphics[width=\linewidth]{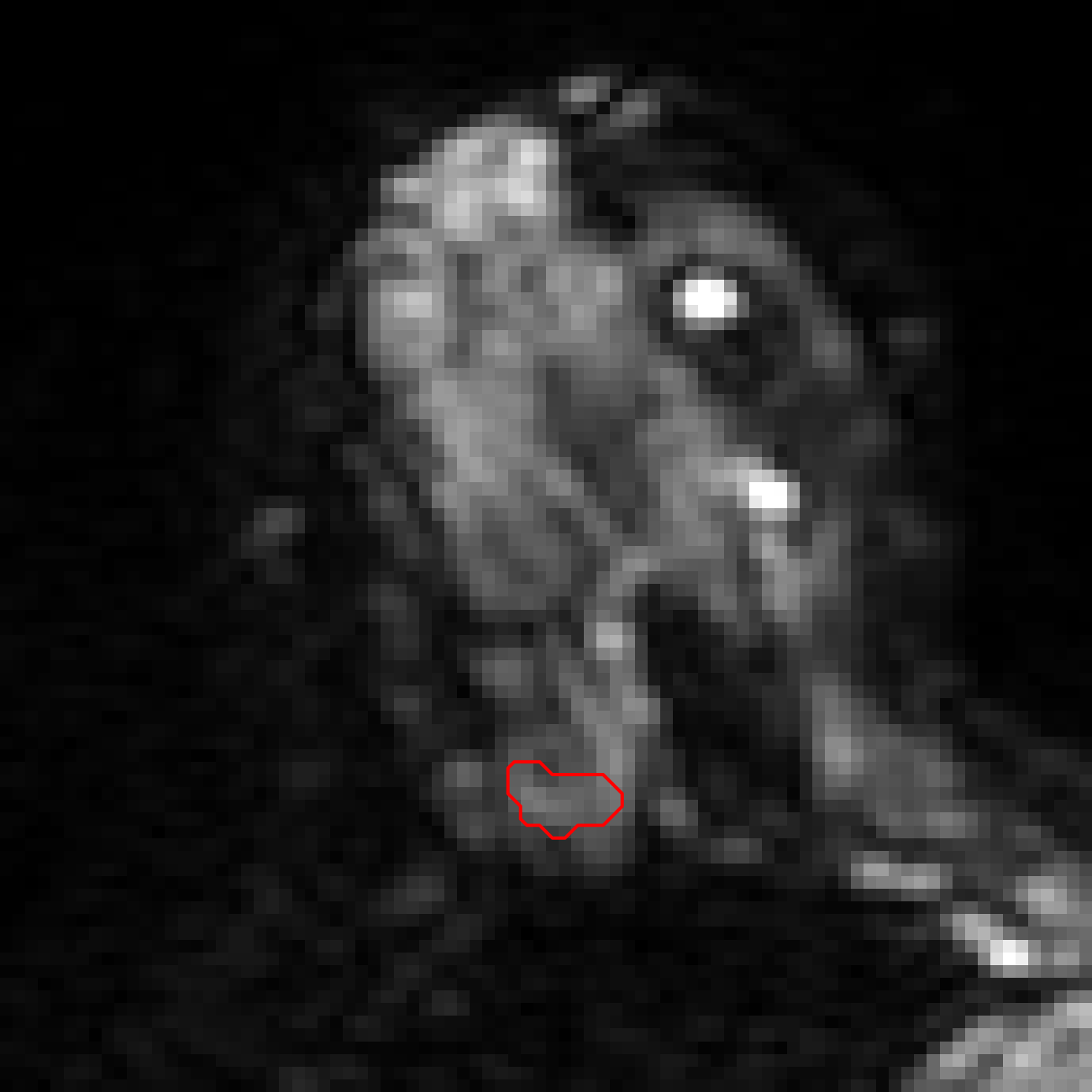}
    \end{subfigure}
    \hfill
    \begin{subfigure}{0.22\textwidth}
        \includegraphics[width=\linewidth]{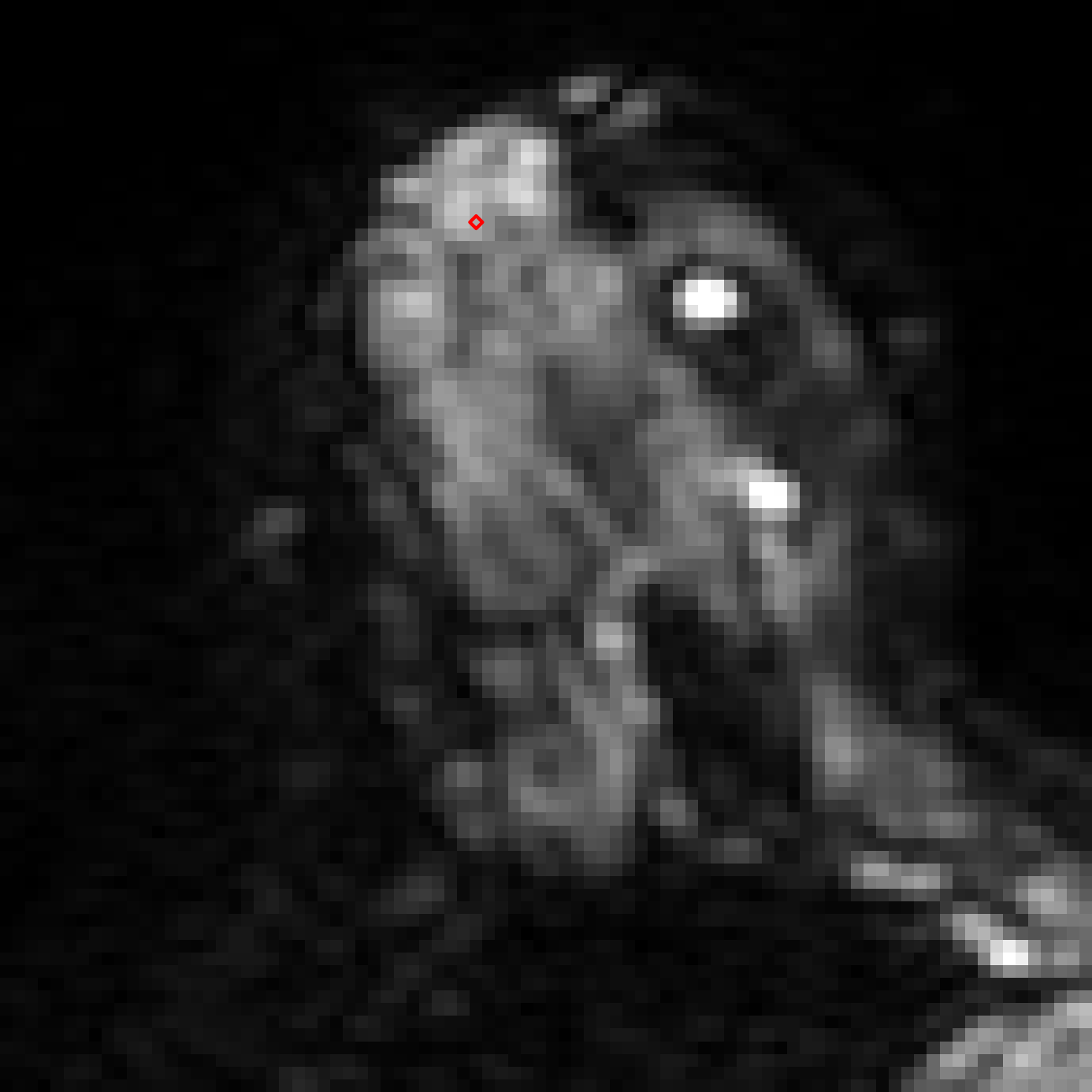}
    \end{subfigure}
  \hfill
    \begin{subfigure}{0.22\textwidth}
        \includegraphics[width=\linewidth]{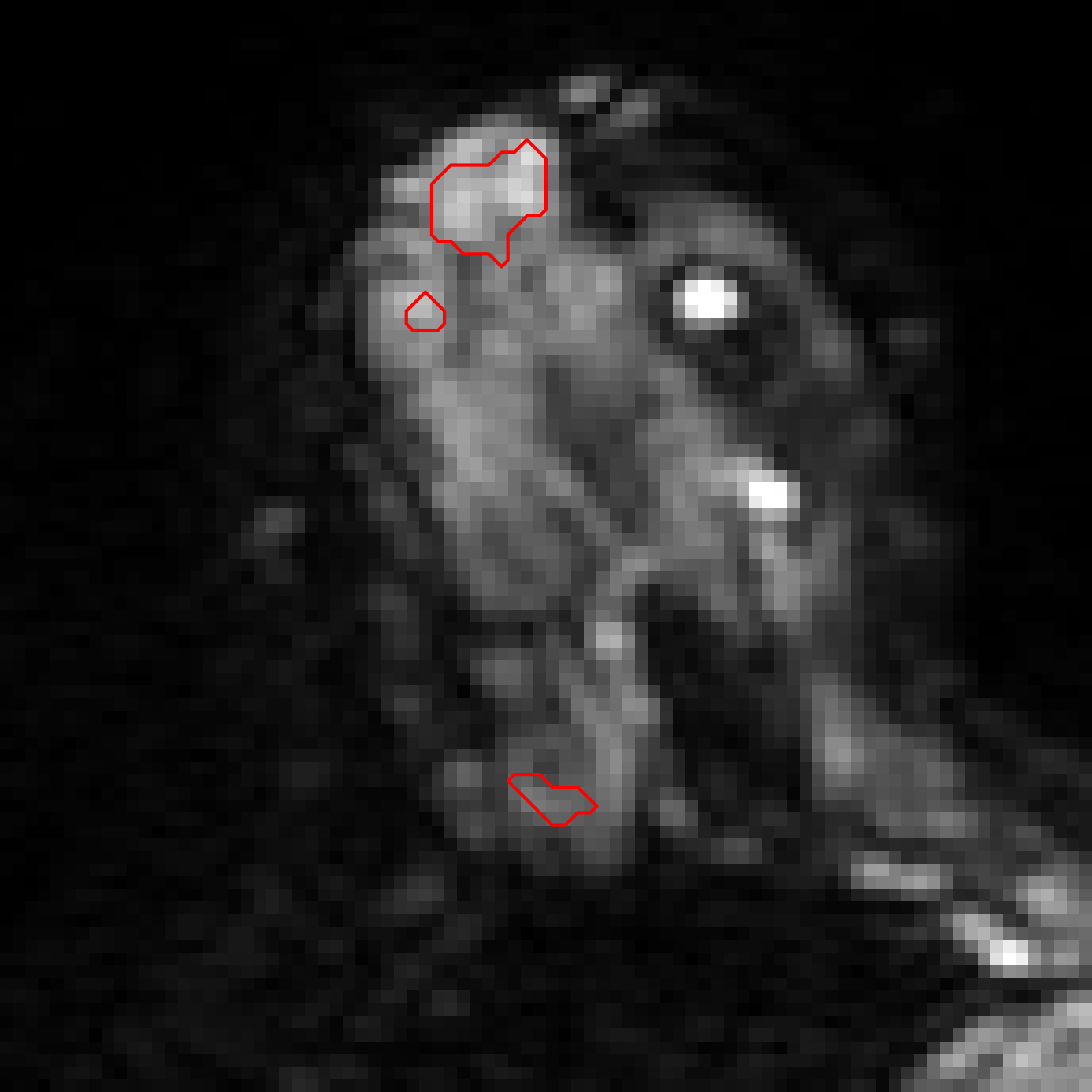}
    \end{subfigure}

    \vspace{2mm} % Some vertical space between the two rows

    % third Row
    \begin{subfigure}{0.22\textwidth}
        \includegraphics[width=\linewidth]{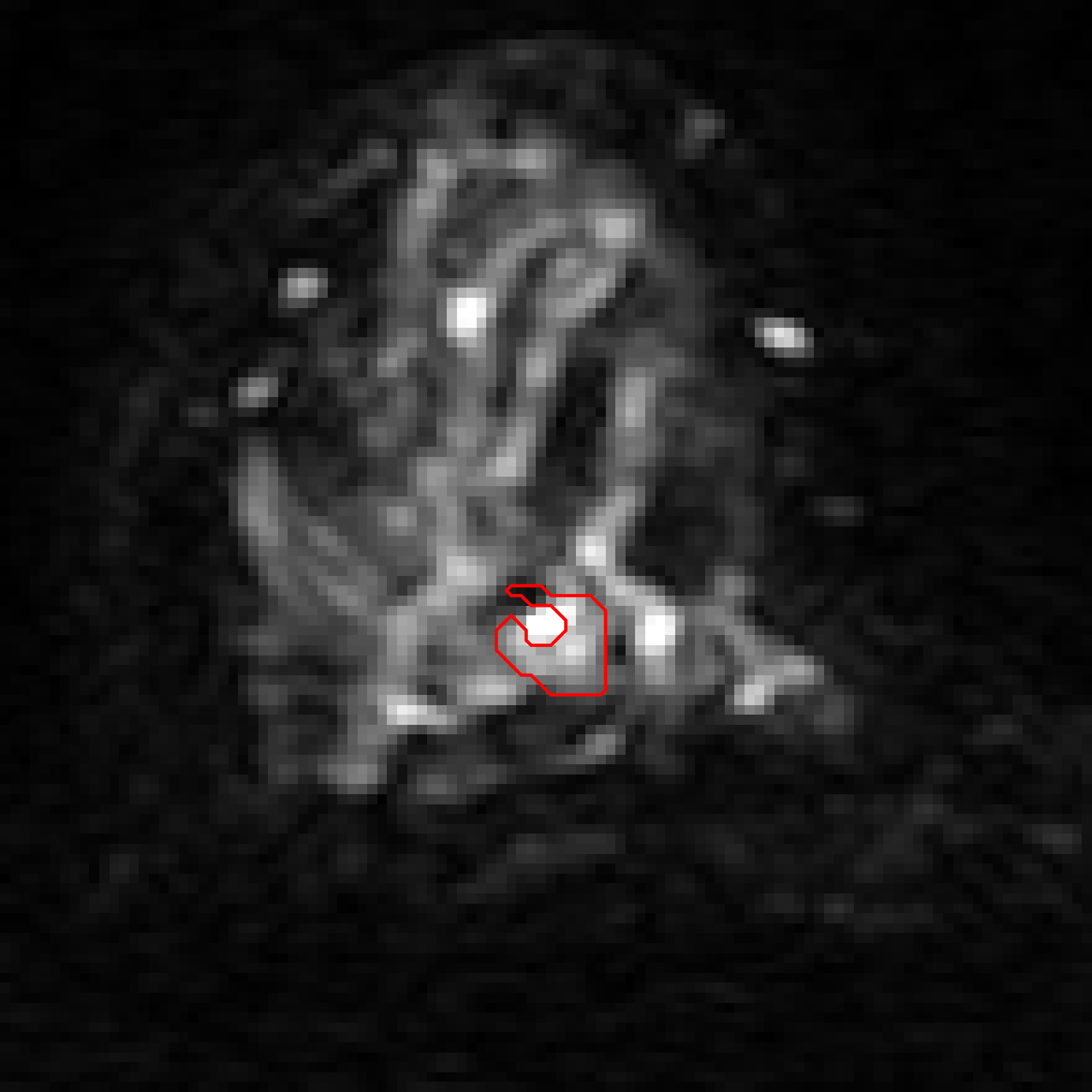}
    \end{subfigure}
    \hfill
    \begin{subfigure}{0.22\textwidth}
        \includegraphics[width=\linewidth]{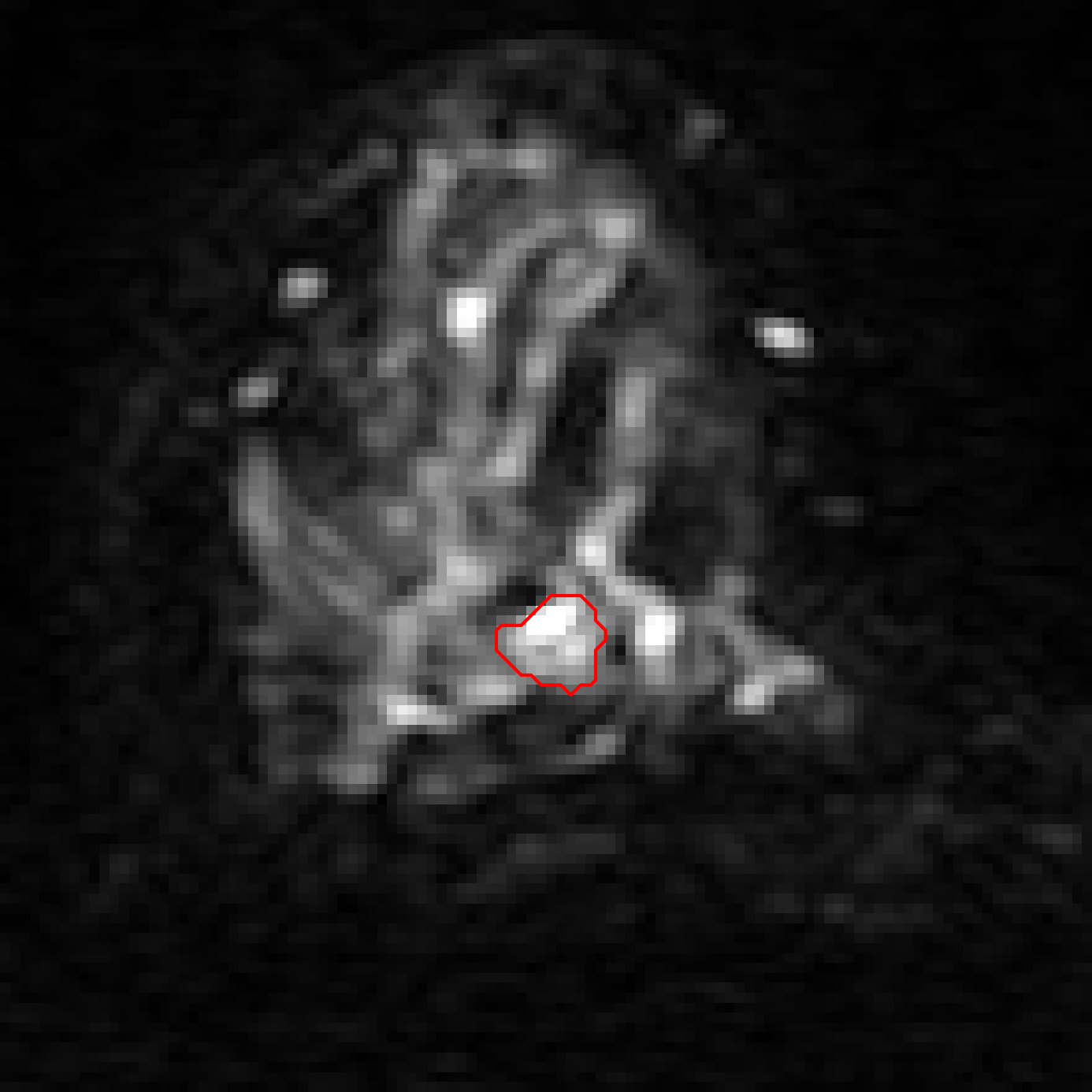}
    \end{subfigure}
    \hfill
    \begin{subfigure}{0.22\textwidth}
        \includegraphics[width=\linewidth]{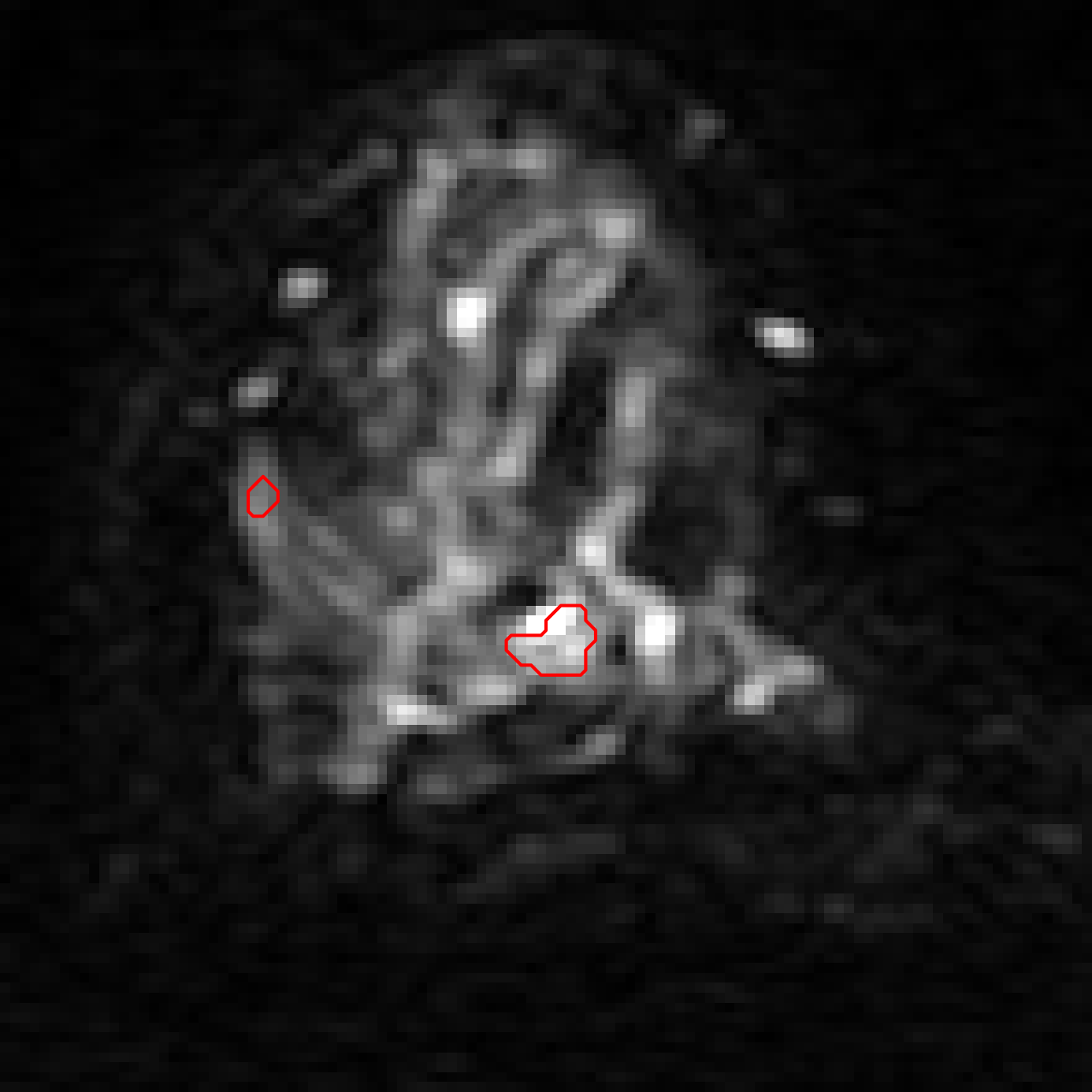}
    \end{subfigure}
    \hfill
    \begin{subfigure}{0.22\textwidth}
        \includegraphics[width=\linewidth]{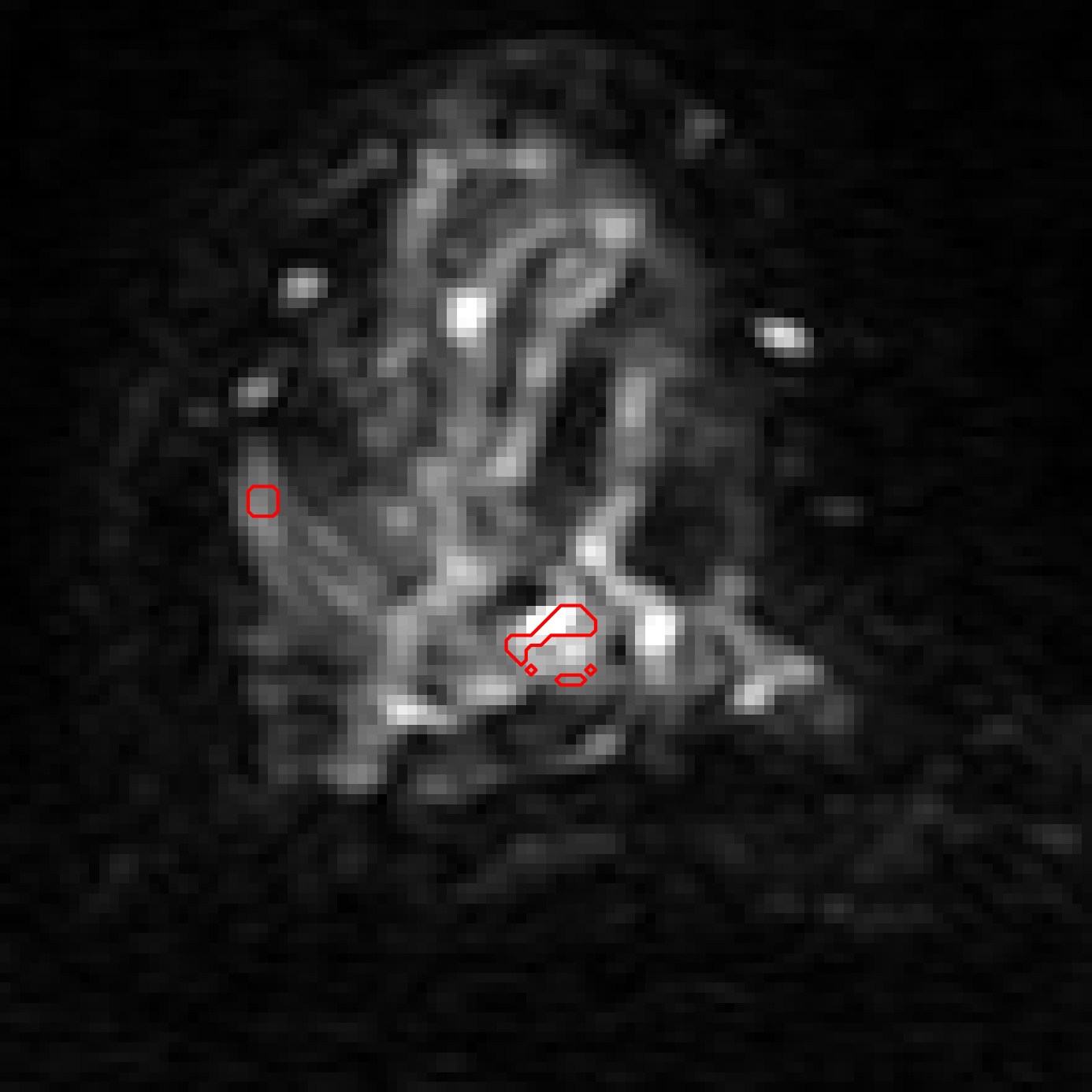}
    \end{subfigure}

    % \vspace{1em} % Some vertical space between the two rows

    % \label{fig:dwi_images}
    %     % forth Row
    % \begin{subfigure}{0.24\textwidth}
    %     \includegraphics[width=\linewidth]{./segmentations/gt_156.png}
    % \end{subfigure}
    % \hfill
    % \begin{subfigure}{0.24\textwidth}
    %     \includegraphics[width=\linewidth]{./segmentations/predicted_04_leision_weight_156.png}
    % \end{subfigure}
    % \hfill
    % \begin{subfigure}{0.24\textwidth}
    %     \includegraphics[width=\linewidth]{./segmentations/CeTverskynifti_156.png}
    % \end{subfigure}
    % \hfill
    % \begin{subfigure}{0.24\textwidth}
    %     \includegraphics[width=\linewidth]{./segmentations/tversky_2k_epochs_156.png}
    % \end{subfigure}

    % \vspace{1em} % Some vertical space between the two rows
        \caption{DWI images with the reference manual and models segmentations.}
    \label{fig:dwi_images}
\end{figure*}

To optimize performance, we combine the WLT and CE losses into a unified loss function:
\begin{equation}
\text{Loss}_{\text{combined}} = \lambda \cdot \text{Loss}_{\text{CE}}(P, \hat{P}) + (1-\lambda) \cdot \text{Loss}_{\text{WLT}}(P, \hat{P}, \Omega)
\end{equation}

The parameter \( \lambda \) serves to balance the contributions of WLT and CE losses, guiding the model to precisely identify smaller lesions and fostering balanced segmentation performance.

Utilizing the nnU-Net combined with our custom-defined loss function, we produced a predicted segmentation \( S_{i}^{\text{pred}} \) for each patient based on their DWI image series \( DWI = \{ I_{i}^{b} \}_{i=1}^{N} \). Here, \( N \) represents the total number of patients, while \( I_{i}^{b} \) indicates the DWI image for the \( i \)-th patient with b-value \( b \in \{0, 100, 600, 800\} \).

\subsubsection{PD-DWI}
We employed the PD-DWI model \cite{gilad2022pd} for pCR prediction, leveraging its proven potential in predicting pCR response in invasive breast cancer cases.
 Using \( S_{i}^{\text{pred}} \) and \( I_{i}^{b} \) as inputs, we predicted pCR, denoted by \( P_{i}^{\text{pCR}} \):
\begin{equation}
P_{i}^{\text{pCR}} = \text{PD-DWI}(S_{i}^{\text{pred}}, I_{i}^{b})    
\end{equation}

% This strategy encompasses three steps visualized in Fig. \ref{workflow}.
\section{Experimental Methodology}
% All the input images of DWI were unified 
\subsection{Training Details}
The model was trained using the nnU-Net framework on the BMMR2 training set. 
Specifically, for our input to the nnU-Net, we utilized DWI images with b-values of 0, 100, 600, and 800. 
 We adopted a 5-fold cross-validation strategy with a batch size of 2.
% For the Tversky loss parameters, the optimal results achieved with \(\alpha = 0.3\) and \(\beta = 0.7\).
In the size-adapting lesion weight loss, the optimal parameters were: \(w_{\text{max}} = 10\), \(w_{\text{min}} = 1\), \(\text{vrange} = 350\), \(k = 7\) and \(\alpha = \sqrt{e^7}\).
To place greater emphasis on the Weighted Lesion Loss for false negatives, we set the Tversky loss parameters to \(\alpha = 0.3\) and \(\beta = 1\).

\subsection{Performance Evaluation}
To evaluate the quality of our segmentations, we utilized the Dice similarity coefficient (DSC) and Hausdorff distance (HD), taking into account data imbalance and the presence of small lesions. For the assessment of pCR prediction performance, metrics such as the Area Under the Curve (AUC) and Cohen's Kappa (Kappa) were employed.

Out of the dataset, 74 cases were used for testing as predefined by the BMMR2 challenge organizers. Among these, 31.08\% were identified as pCR and 68.92\% as non-pCR. Cases that the segmentation model predicted with empty segmentation were categorized as non-pCR, in line with this distribution.

Our method's pCR prediction performance, which uses our automated segmentation with the Weighted Lesion approach, was then compared against results obtained from both manual segmentation and automated segmentation without the Weighted Lesion enhancement.

\section{Results}
\subsection{Performance at pre-NAC}
Table \ref{table:table_t0} summarizes the performance of various methods at the pre-NAC stage. Our nnU-Net model with Tversky loss demonstrates human-level sensitivity and outperforms the manual method in terms of specificity and Kappa. When we incorporated the size-adaptive lesion weighting loss, the AUC showed a notable improvement compared to automated segmentation methods without this enhancement.

% At the initial timepoint, T0 (pre-NAC), the lesions are generally larger, making the distinctions introduced by our "Size-Adaptive Lesion Weighting" loss less pronounced. It is essential to highlight that the inherent strength of our proposed method becomes significantly more evident in the subsequent stages of treatment. 
% % As depicted in Figure, the mean size of tumors decreases across timepoints.
% By the time we reach T2, the tumors not only become substantially smaller but also exhibit a pronounced variability in size.

\begin{table}[b]
\caption{Metrics at pre-NAC, with T0 data}
\label{table:table_t0}
\begin{center}
\begin{tabular}{|c||c|c|c|c|c|c|}
\hline
& Kappa & AUC & Dice & HD \\
\hline
Manual & 0.458 & 0.796 & - & - \\
\hline
Lesion Weighting 
 & 0.330 & 0.760 & 0.590 & 48.16 \\
\hline
Tversky & 0.507 & 0.751 & 0.584 & 97.50 \\
\hline

Tversky with CE
 & 0.277 & 0.676 & 0.573 & 60.21 \\
\hline

\end{tabular}
\end{center}
\end{table}

\subsection{Performance during NAC treatment}

\begin{figure}[t!]
    \centering
    \includegraphics[width=0.48\textwidth]{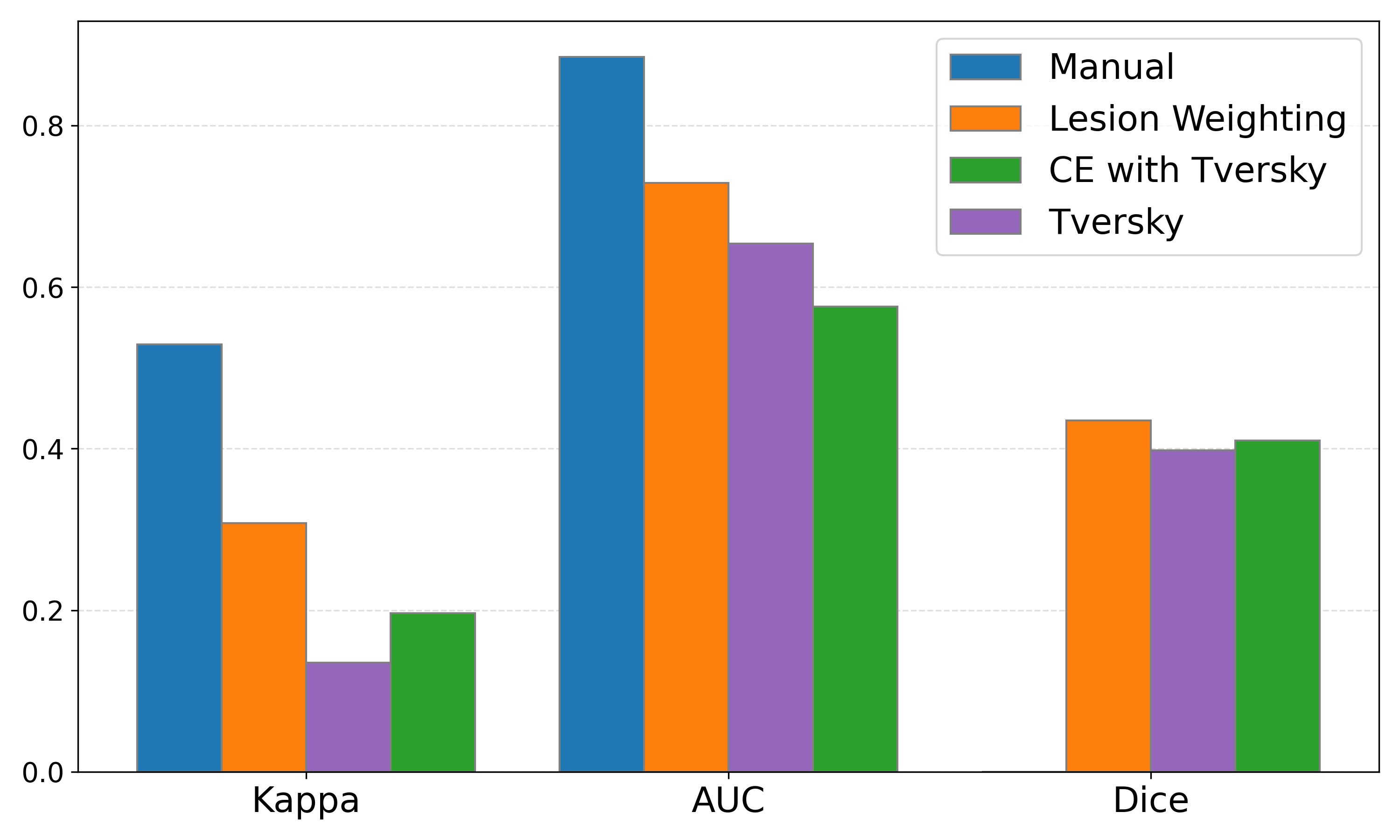}
    \caption{Metrics at mid-NAC, with T0, T1, and T2 data}
    \label{bar_chart_t2}
\end{figure}
The assessment of segmentation quality during NAC treatment is crucial as it reflects the changing characteristics of the tumor in response to the therapy. Figure ~\ref{fig:dwi_images} illustrates the DWI images with reference manual and models segmentations at this stage, providing a clear visual representation of the effectiveness of our segmentation approach during NAC treatment.

Figure ~\ref{bar_chart_t2} provides an overview of the performance metrics of various methods at the mid-NAC stage, drawing on data from T0, T1, and T2 timepoints. Our size-adaptive lesion weighting loss outperformed other automated segmentation methods that did not incorporate this technique, across all metrics. 

% Figure ~\ref{bar_chart_t2} provides an overview of the performance metrics of various methods at Mid-NAC treatment, relying on data from timepoints T0, T1, and T2. Our size-adaptive lesion weighting loss outperformed other automated segmentation methods that did not incorporate this technique, across all metrics. 

% This is visually evidenced in the ROC curves for the T1 and T2 timepoints, as depicted in Figures \ref{fig:roc_t1} and \ref{fig:roc_t2}.

The significant improvement in results during these complex stages of NAC treatment, where the tumor often shrinks considerably and displays increased variability in size and shape, underscores the advantage of employing the size-adaptive lesion weight loss at these stages.

\section{CONCLUSION}
We presented the ``Size-Adaptive Lesion Weighting'' loss incorporated into the nnU-Net architecture, a pioneering methodology that prioritizes lesion significance during training based on their size. By bridging the gap between precise tumor segmentation and prediction, our approach not only optimizes the accuracy of pCR predictions but also eradicates the need for manual tumor segmentation, effectively streamlining the pCR prediction process.

\section{Compliance with ethical standards}
\label{sec:ethics}

  This research study was conducted retrospectively using human subject data made available in open access by \cite{partridge2018diffusion}. Ethical approval was not required as confirmed by the license attached with the open-access data.

\section{Conflict of Interest declaration}
The authors have no relevant financial or non-financial interests to disclose.

\section{Acknowledgments}
\label{sec:acknowledgments}
This work was supported in part by research grants from the Israel-US Binational Science Foundation, the Israeli Ministry of Science and Technology, the Israel Innovation Authority, and the joint Microsoft Education and the Israel Inter-university Computation Center (IUCC) program.

\bibliographystyle{IEEEbib}
\bibliography{refs}

\end{document}